\documentclass[11pt]{article}
\usepackage{amsmath}
\usepackage[final]{acl}

\usepackage{times}
\usepackage{latexsym}
\usepackage{multirow}
\usepackage[T1]{fontenc}

\usepackage[utf8]{inputenc}

\usepackage{microtype}

\usepackage{inconsolata}

\usepackage{graphicx}

\title{Detecting AI-Generated Content in Academic Peer Reviews}

\author{
Siyuan Shen \quad Kai Wang
\\[0.5ex]
University of Pennsylvania, Philadelphia, PA, USA
\\
Children’s Hospital of Philadelphia, Philadelphia, PA, USA
\\
shsiyuan@sas.upenn.edu, wangk@chop.edu
\\[1ex]
}

\begin{document}
\maketitle
\begin{abstract}
The growing availability of large language models (LLMs) has raised questions about their role in academic peer review. This study examines the temporal emergence of AI-generated content in peer reviews by applying a detection model trained on historical reviews to later review cycles at \textit{International Conference on Learning Representations} (ICLR) and \textit{Nature Communications} (NC). We observe minimal detection of AI-generated content before 2022, followed by a substantial increase through 2025, with approximately 20\% of ICLR reviews and 12\% of Nature Communications reviews classified as AI-generated in 2025. The most pronounced growth of AI-generated reviews in NC occurs between the third and fourth quarter of 2024. Together, these findings provide suggestive evidence of a rapidly increasing presence of AI-assisted content in peer review and highlight the need for further study of its implications for scholarly evaluation.
\end{abstract}

\section{Introduction}
\label{sec:introduction}

Recent advances in large language models (LLMs) have substantially changed how text is produced across many domains, including scientific writing. Alongside their benefits, these models have raised concerns about their potential influence on the academic peer review process. In particular, the use of AI tools to generate or assist in writing peer reviews may affect the integrity, reliability, and transparency of scholarly evaluation. Notably, some major conferences, such as ICML 2025, have begun to incorporate AI reviewers into the review process, although authors are typically not required to address comments generated by AI reviewers. Despite growing discussion around this issue, empirical evidence on how AI-generated content has emerged and evolved in peer reviews remains limited.

A key challenge in studying AI use in peer review is the lack of direct disclosure.
Reviews are typically anonymous, and reviewers are not required to report whether AI tools were used.
As a result, understanding the prevalence of AI-generated or AI-assisted reviews requires indirect approaches that can operate on large collections of historical data.
Moreover, since language models continue to evolve rapidly, it is important to assess not only whether AI-generated reviews exist, but also how their presence changes over time.

In this work, we present a systematic, data-driven analysis of AI-generated content in academic peer reviews.
We adopt a detection-based framework in which a classifier is trained to distinguish between human-written and AI-generated reviews using labeled data from a historical year.
The trained detector is then applied to reviews from subsequent years, enabling an analysis of temporal generalization under a realistic deployment setting.
This design allows us to study longitudinal trends while holding the detection model fixed.

We conduct experiments on peer reviews from two representative publication venues: \textit{International Conference on Learning Representations} (ICLR), which follows an annual conference-based review cycle, and \textit{Nature Communications} (NC), which operates under a continuous journal review process. The combination of these venues enables analysis across different review formats and temporal granularities, including both yearly and quarterly trends. Across both settings, we observe consistent temporal patterns in reviews classified as AI-generated, suggesting that the phenomenon is not specific to a single venue or review structure.

\section{Literature Review}
Recent studies and reports have begun to examine the implications of large language models for the integrity of academic peer review. A comprehensive perspective is provided by the work titled ``Ensuring peer review integrity in the era of large language models: A critical stocktaking of challenges, red flags, and recommendations,'' which systematically analyzes the risks posed by LLM use in peer review~\cite{peerreview_llm_integrity}. The authors argue that while language models may offer benefits such as improved efficiency and linguistic clarity, their undisclosed or unregulated use can undermine transparency, blur the boundary between human judgment and automated assistance, and complicate editorial oversight.

Complementing conceptual analyses, empirical and observational evidence has also begun to emerge. Liang et al.\ analyze large-scale conference review data and estimate the prevalence of AI-modified and AI-generated content in peer reviews, providing one of the first quantitative measurements of this phenomenon~\cite{liang2024chatgpt_reviews}. The main findings are that 10.6\% of ICLR 2024 review sentences and 16.9\% for EMNLP have been substantially modified by ChatGPT, with no significant evidence
of ChatGPT usage in Nature portfolio reviews. In addition, a News article published in Nature reported that a survey of 1,600 academics showed that more than 50\% have used artificial-intelligence tools while peer reviewing manuscripts, even when the core judgments remain human-authored~\cite{nature_ai_assist_reviews}. 

Beyond scholarly analysis, real-world incidents further underscore the urgency of these concerns. A News article published in Nature documents that a major artificial intelligence conference received a substantial number of peer reviews that were fully generated by AI systems, with many additional reviews showing signs of AI assistance~\cite{nature_ai_conference_flooded}. Specifically, analysis by Pangram Labs revealed that around 21\% of the ICLR peer reviews were fully AI-generated, and more than half contained signs of AI use. This case illustrates that AI-generated peer reviews are not merely a hypothetical risk but an emerging phenomenon in practice, raising questions about review quality, accountability, and the scalability of editorial oversight.

Together, these works suggest both the conceptual challenges and empirical realities of AI use in peer review, motivating systematic, data-driven investigation into the prevalence and temporal dynamics of AI-generated content in scholarly evaluation.

\section{Experimental Design}
\label{sec:experimental_design}

\subsection{Task}
\label{sec:task}

We formulate the problem as a binary classification task that aims to determine whether a given peer review is written by a human reviewer or generated by an AI system. In practice, some reviewers may draft their reviews manually and subsequently polish or revise the text using AI tools, resulting in hybrid human–AI content. Each input consists of the full textual content of a single peer review, and the output is a binary label indicating whether the review is real or AI-generated.

To examine the temporal generalization of detection models, we adopt a year-based evaluation protocol.
Models are trained exclusively on reviews from a single historical year and evaluated on reviews from subsequent years.
This design reflects a realistic deployment scenario in which a detector trained on past data is applied to future review submissions whose writing characteristics may change over time.

\subsection{Data}
\label{sec:data}

\subsubsection{Real Reviews}
\label{sec:real_reviews}

We collect real peer reviews from two publication venues: ICLR and NC.
ICLR reviews are produced within a single annual conference review cycle, yielding one fixed snapshot of reviews per year.
In contrast, NC follows an ongoing journal review process, where reviews are continuously submitted and updated throughout the year.
Accordingly, we aggregate NC reviews by calendar year for the main analysis and further partition them into quarters for finer-grained temporal analysis.

To ensure comparability across years while controlling data scale, we adopt a paper-level sampling strategy.
For each evaluation year, we randomly sample approximately 500 papers and include all associated reviews, resulting in roughly 2,000 review texts per year.
This approach preserves reviewer diversity and mitigates bias introduced by paper-level heterogeneity.

Reviews from the year 2021 are used exclusively for training, while reviews from 2022 to 2025 are reserved for evaluation.
No review text appears in more than one data split.

\subsubsection{Synthetic Reviews}
\label{sec:synthetic_reviews}

To construct AI-generated reviews, we employ large language models to produce synthetic peer review texts conditioned on paper content.
Specifically, synthetic reviews are generated using the DeepSeek Reasoner API.
The generation process is designed to approximate realistic reviewer behavior in terms of tone, structure, and length, while avoiding trivial lexical artifacts.

Synthetic reviews are generated to align with the distribution of real reviews from the corresponding training year.
This alignment includes constraints on review length and discourse structure, ensuring that the detection task focuses on stylistic and semantic signals rather than superficial differences.
Details of the prompting strategy and API configuration are provided in the appendix.

\subsection{Detection Model}
\label{sec:detection_model}

\subsubsection{Model Architecture}
\label{sec:model_architecture}
Taking ICLR 2025 reviews as an illustrative example, the average review length is approximately 447 tokens, with the longest reviews exceeding 2,000 tokens. Such lengths substantially exceed the practical input limits of standard transformer models with full self-attention. Accordingly, we adopt Longformer as the backbone architecture for review classification~\citep{beltagy2020longformer}, as it is specifically designed to handle long-form documents through sparse attention mechanisms and efficient long-sequence modeling.

To enable parameter-efficient fine-tuning and reduce overfitting on limited training data, we apply Low-Rank Adaptation to the backbone model~\citep{hu2022lora}.
Only a small number of additional parameters are trained, while the pretrained model weights remain frozen.

\subsubsection{Training Strategy}
\label{sec:training_strategy}

Models are trained on labeled reviews from the 2021 dataset and evaluated on reviews from later years without further adaptation. We use a balanced training set consisting of equal numbers of real and synthetic reviews to prevent class imbalance from influencing the decision boundary. In addition, we provide representative examples of human-written and AI-generated reviews in the supplementary materials to illustrate their linguistic differences and facilitate intuitive understanding.

All models are trained using standard cross-entropy loss.
Hyperparameters, including learning rate and batch size, are fixed across all experiments to ensure consistency and comparability.
The full set of training hyperparameters and implementation details is provided in the appendix. Reproducible code is available at https://github.com/sy-shen/AI\_Review.

\section{Results}
\label{sec:results}

\subsection{ICLR Results}
\label{sec:iclr_results}

The detection model is trained on ICLR 2021 reviews and evaluated on reviews from 2022 to 2025.

\subsubsection{Model Performance on Training Set (ICLR 2021)}
\label{sec:iclr_training_performance}

The detection model achieves perfect classification performance on ICLR reviews from 2021.
Specifically, all 160 real reviews are predicted as real, and all 160 AI-generated reviews are predicted as AI.

\subsubsection{Model Inference Results (ICLR 2022--2025)}
\label{sec:iclr_trends}

\begin{figure*}[t]
\centering
\includegraphics[width=0.9\textwidth]{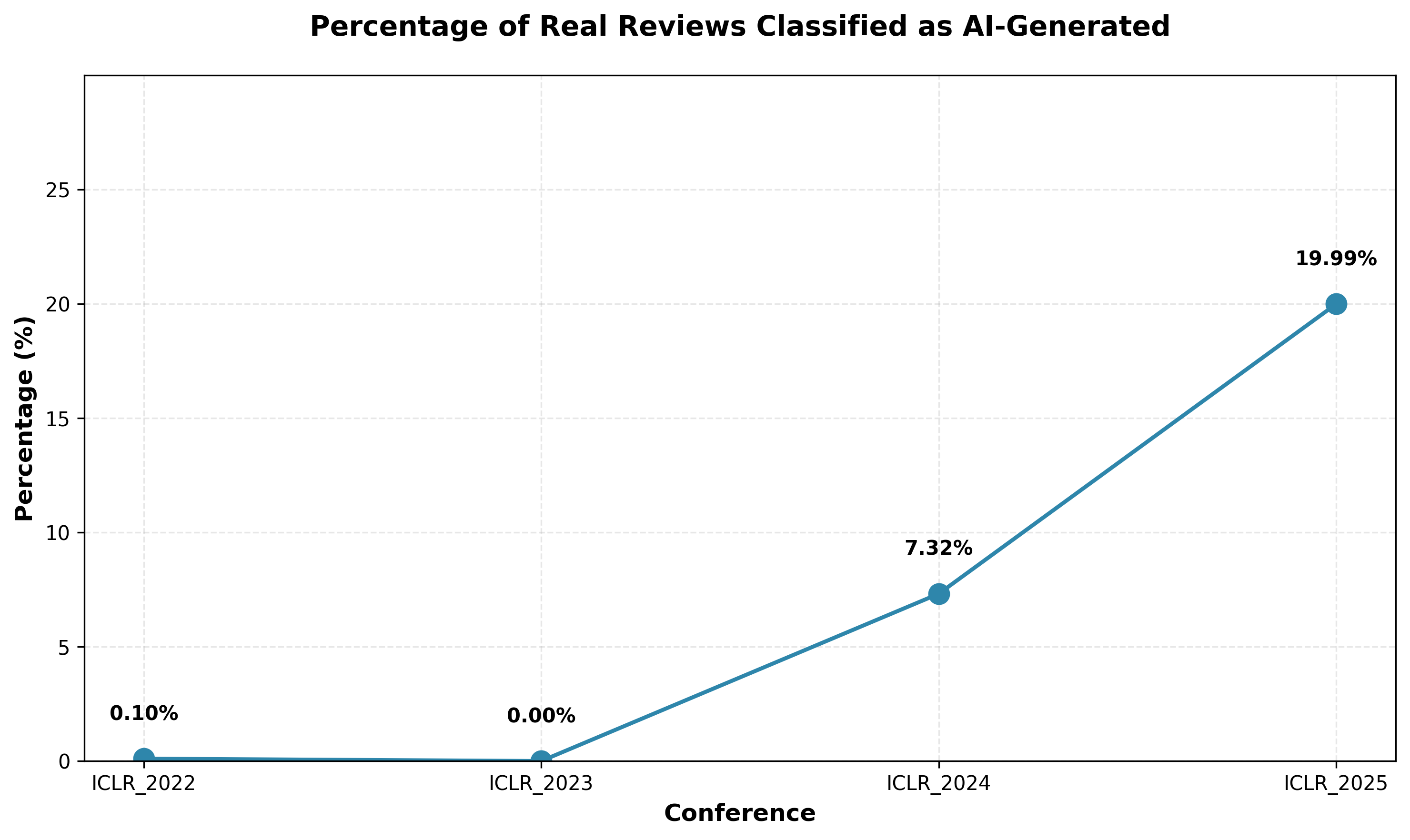}
\caption{Percentage of ICLR reviews classified as AI-generated from 2022 to 2025.}
\label{fig:iclr_ai_percentage_trend}
\end{figure*}

Figure~\ref{fig:iclr_ai_percentage_trend} presents the percentage of ICLR reviews classified as AI-generated for each evaluation year from 2022 to 2025.
Table~\ref{tab:iclr_yearly_stats} reports the corresponding counts and percentages.
The proportion of reviews classified as AI-generated increases across successive evaluation years.

\begin{table}[t]
\centering
\resizebox{\columnwidth}{!}{%
\begin{tabular}{l r r r}
\hline
Year & Total Reviews & AI-Detected & Percentage (\%) \\
\hline
2022 & 1,937 & 2   & 0.10 \\
2023 & 1,887 & 0   & 0.00 \\
2024 & 1,818 & 133 & 7.32 \\
2025 & 1,961 & 392 & 19.99 \\
\hline
\end{tabular}%
}
\caption{Summary statistics of prediction of AI-generated review for ICLR.}
\label{tab:iclr_yearly_stats}
\end{table}

\subsection{NC Results}
\label{sec:nc_results}

The detection model is trained on NC 2021 reviews and evaluated on reviews from 2022 to 2025.

\subsubsection{Model Performance on Training Set (NC 2021)}
\label{sec:nc_training_performance}

The detection model achieves perfect classification performance on NC reviews from 2021.
Specifically, all 120 real reviews are predicted as real, and all 120 AI-generated reviews are predicted as AI.

\subsubsection{Model Inference Results by Year (NC 2022--2025)}
\label{sec:nc_yearly_trends}

\begin{figure*}[t]
\centering
\includegraphics[width=0.9\linewidth]{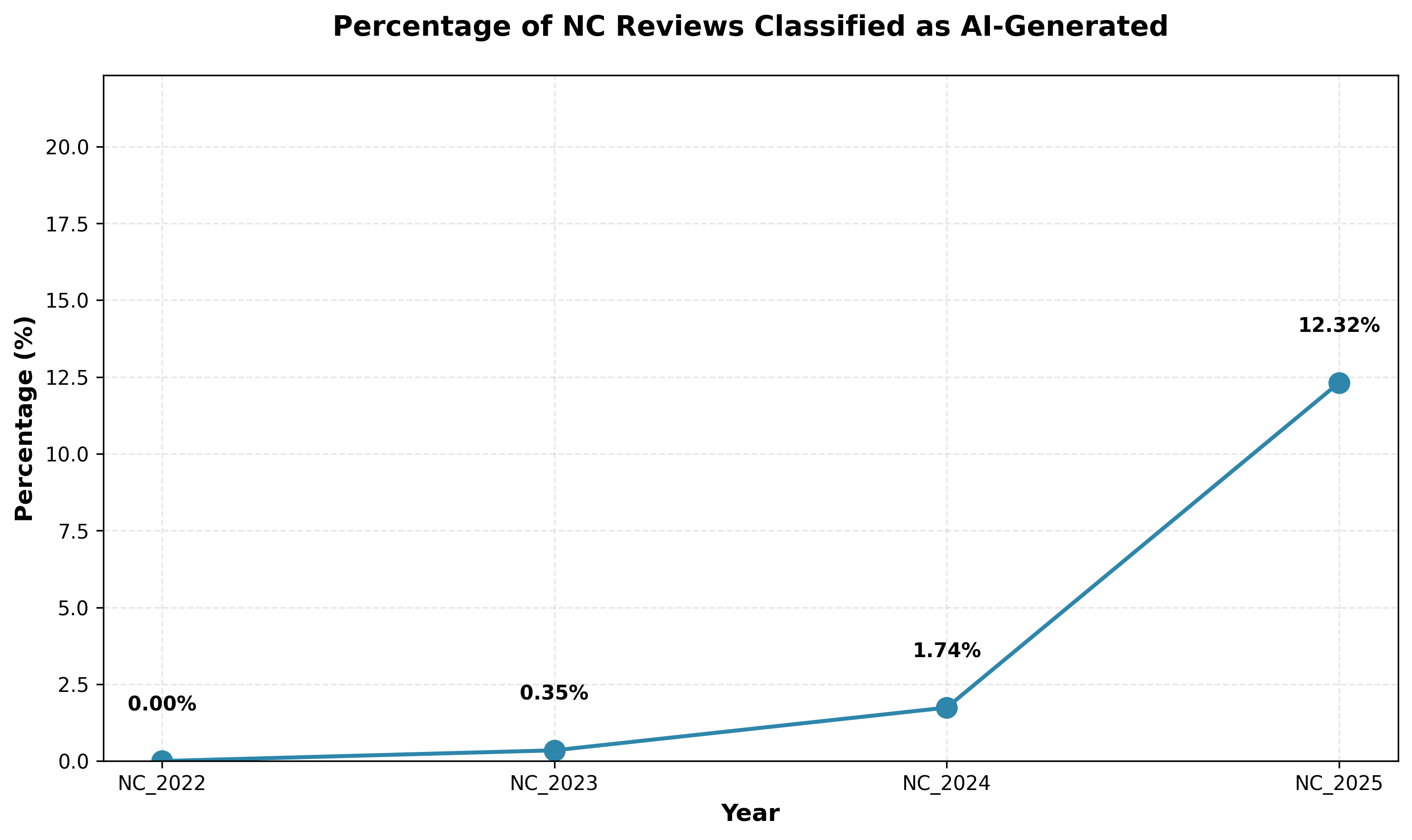}
\caption{Percentage of NC reviews classified as AI-generated from 2022 to 2025.}
\label{fig:nc_ai_percentage_trend}
\end{figure*}

Figure~\ref{fig:nc_ai_percentage_trend} presents the yearly percentage of NC reviews classified as AI-generated from 2022 to 2025.
Table~\ref{tab:nc_yearly_percentage} reports the corresponding yearly percentages.
The proportion of reviews classified as AI-generated increases across successive evaluation years.

\begin{table}[t]
\centering
\resizebox{\columnwidth}{!}{%
\begin{tabular}{l r r r}
\hline
Year & Total Reviews & AI-Detected & Percentage (\%) \\
\hline
2022 & 348 & 0  & 0.00 \\
2023 & 287 & 1  & 0.35 \\
2024 & 460 & 8  & 1.74 \\
2025 & 406 & 50 & 12.32 \\
\hline
\end{tabular}%
}
\caption{Summary statistics of prediction of AI-generated review for NC.}
\label{tab:nc_yearly_percentage}
\end{table}

\subsubsection{Quarterly Inference Results (NC)}
\label{sec:nc_quarterly_trends}

\begin{figure*}[t]
\centering
\includegraphics[width=0.9\linewidth]{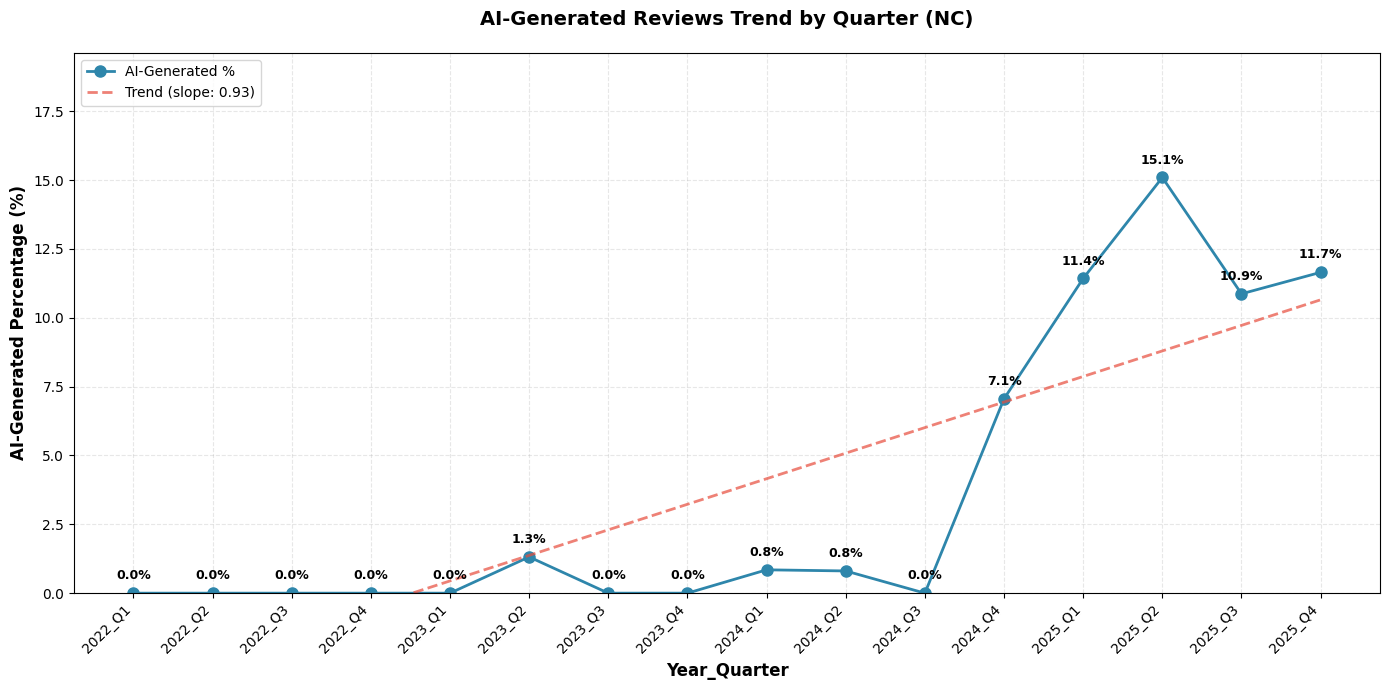}
\caption{Quarterly percentage of NC reviews classified as AI-generated.}
\label{fig:nc_ai_percentage_trend_quarterly}
\end{figure*}

Figure~\ref{fig:nc_ai_percentage_trend_quarterly} presents the quarterly percentage of NC reviews classified as AI-generated.
An increasing proportion of reviews is observed across successive quarters. The most pronounced growth in NC occurs between the third and fourth quarter of 2024.

\section{Discussion}
\label{sec:discussion}

Across both ICLR and NC, the experimental results reveal a consistent temporal pattern in the detection of AI-generated peer reviews.
In earlier evaluation years, the detection model identifies almost no AI-generated reviews, whereas substantially higher proportions are observed in later years.

For ICLR, reviews from 2022 and 2023 contain virtually no content classified as AI-generated.
This observation aligns with the timing of large-scale deployment of conversational language models and suggests that AI-assisted review writing was uncommon during earlier conference review cycles.
The sharp increase observed from 2024 onward indicates a notable shift in review-writing practices within a relatively short time span.

A similar pattern is observed for NC.
Year-level results show minimal detection rates in 2022 and 2023, followed by a steady increase in 2024 and a pronounced rise in 2025. We note that an earlier study in 2024 ~\cite{liang2024chatgpt_reviews} found no significant evidence
of ChatGPT usage in Nature portfolio reviews, but this may be explained by the inclusion of only pre-2024 NC reviews.
The availability of finer-grained temporal data for NC further enables a quarterly analysis, which reveals a gradual upward trajectory rather than an abrupt change. The most notable increase occurs between the third and fourth quarter of 2024. As ChatGPT-4o was officially released by OpenAI in May 2024 to be available to free-tier users, with multimodal input capabilities, it may partially explain the observed trend.
In summary, our findings suggest that the adoption of AI tools in journal peer review may have increased progressively over time, though lags slightly behind machine learning conferences.

Despite differences between conference-based and journal-based review processes, the parallel trends observed in ICLR and NC point to a broader phenomenon that is not specific to a single venue or review format.
The consistency across venues strengthens the interpretation that the observed increase reflects a genuine change in reviewer behavior, rather than artifacts of a particular dataset or evaluation setting.

Our study has several important limitations. First, the sample size is relatively limited, as we analyze only a subset of reviews from ICLR and Nature Communications. Nevertheless, the temporal trends we observe are pronounced and consistent with prior empirical findings, suggesting that the signals captured by our analysis are robust. Second, our formulation of the task as a binary classification problem to distinguishing between “AI-generated” and “human-written” certainly over-simplifies the underlying phenomenon. In practice, many peer reviews (as suggested by the survey~\cite{nature_ai_assist_reviews}) likely reflect combined human and AI efforts, where core scientific judgments are written by humans and subsequently polished by AI. Detecting such partial AI assistance at the sentence or paragraph level remains challenging, and the reliability of fine-grained attribution is currently unclear. Therefore, we adopt a binary classification strategy in the current study. As a result, the reported percentages should be interpreted as estimates of reviews that exhibit strong AI-generated characteristics with minimal human editing, and they likely underestimate the prevalence of AI-assisted review writing. Third, the study could be strengthened through additional robustness analyses. Future work could extend this approach to journals from other disciplines (such as biology, medicine, chemistry and social sciences), explore multiple prompting strategies for synthetic data generation, and incorporate surveys or annotations from human reviewers to better characterize real-world patterns of AI use in peer review. Finally, we acknowledge the potential risk of overfitting arising from the use of synthetic training data that can be more easily separable than real-world AI-assisted reviews. All synthetic reviews in this study are generated using a single language model (DeepSeek Reasoner) with a fixed prompting strategy, which may not fully capture the diversity of AI usage by academic reviewers. Consequently, it remains unclear how well the detector generalizes to reviews written or edited using other models, such as ChatGPT or Gemini, or to future iterations of the same model family. Addressing these limitations will be an important direction for future research.

Overall, these findings highlight a growing presence of AI-generated or AI-assisted content in academic peer reviews.
They underscore the importance of understanding how emerging language models may influence the peer review process and motivate further investigation into transparency, policy, and best practices surrounding the use of AI tools in scholarly evaluation.
\\

\small{Disclaimer: This study reports the outcome of an independent study of the AMCS 5999 course in the Department of Mathematics at the University of Pennsylvania. The manuscript was written by academic researchers and underwent limited language polishing by AI.}

\bibliography{custom}

\section{Appendix A}
\label{sec:appendix_prompt}

The prompt used to generate synthetic peer reviews for ICLR papers is as follows:

\begin{quote}
As an ICLR reviewer, write a detailed review for this paper.

Write a natural, flowing review that includes a brief summary of what the paper proposes and its main contributions, a discussion of strengths and weaknesses, specific technical questions or concerns, comments on clarity, novelty, and experimental validation, suggestions for improvement, and an overall assessment.

Be professional, constructive, and specific. Write in a natural style similar to real ICLR reviews, rather than using a rigid numbered format.

Write the review in 200--400 words.
\end{quote}

\section{Appendix B}
\label{sec:appendix}
Table~\ref{tab:training_hyperparameters} summarizes the main training hyperparameters used in
our experiments.
\begin{table}[h]
  \centering
  \begin{tabular}{l r}
    \hline
    Hyperparameter & Value \\
    \hline
    Maximum sequence length & 2048 \\
    Batch size & 8 \\
    Number of epochs & 8 \\
    LoRA rank & 8 \\
    \hline
  \end{tabular}
  \caption{Training hyperparameters used for all experiments.}
  \label{tab:training_hyperparameters}
\end{table}

\end{document}